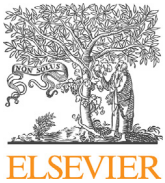

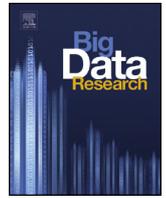

# A Semantic Approach for Big Data Exploration in Industry 4.0

Idoia Berges *, Víctor Julio Ramírez-Durán, Arantza Illarramendi

*University of the Basque Country UPV/EHU, Donostia - San Sebastián, 20018, Spain*

### A B S T R A C T

The growing trends in automation, Internet of Things, big data and cloud computing technologies have led to the fourth industrial revolution (Industry 4.0), where it is possible to visualize and identify patterns and insights, which results in a better understanding of the data and can improve the manufacturing process. However, many times, the task of data exploration results difficult for manufacturing experts because they might be interested in analyzing also data that does not appear in pre-designed visualizations and therefore they must be assisted by Information Technology experts.
In this paper, we present a proposal materialized in a semantic-based visual query system developed for a real Industry 4.0 scenario that allows domain experts to explore and visualize data in a friendly way. The main novelty of the system is the combined use that it makes of captured data that are semantically annotated first, and a 2D customized digital representation of a machine that is also linked with semantic descriptions. Those descriptions are expressed using terms of an ontology, where, among others, the sensors that are used to capture indicators about the performance of a machine that belongs to a Industry 4.0 scenario have been modeled. Moreover, this semantic description allows to: formulate queries at a higher level of abstraction, provide customized graphical visualizations of the results based on the format and nature of the data, and download enriched data enabling further types of analysis.

© 2021 The Authors. Published by Elsevier Inc. This is an open access article under the CC BY-NC-ND license (http://creativecommons.org/licenses/by-nc-nd/4.0/).

## 1. Introduction

We are witnessing a digital transformation era in which ubiquitous sensors and Internet of Things (IoT) devices enable the *datafication* of virtually any aspect of digitally connected individuals and machines [1]. Thus a new kind of economy that is based on data has emerged, coining the concept of data-driven economy [2]. One sector where data-driven economy is being implanted all over the world is the manufacturing industry. The fourth industrial revolution has given rise to what is called Smart Manufacturing, addressing the use of modern Information Technologies (IT) to transform the acquired data into manufacturing intelligence in order to achieve meaningful improvements in all aspects of manufacturing. The term Smart Manufacturing includes different initiatives, among which we can find "Smart Manufacturing" in USA, "Made in CHINA 2025", "Future Manufacturing" in UK and "Industry 4.0" in Europe [3]. These initiatives enable important business opportunities for the manufacturers.

In general, the deployment of Smart Manufacturing approaches demands the introduction of data-related Information Technologies and digital platforms supporting it. Moreover, the design and implementation of such technologies and platforms faces diverse research and innovation challenges. These include, among others, improved methods of gathering valuable machine data and data integration across different sources of heterogeneous nature, implementation of advanced data analytics technologies and methods, and visualizing data to provide the right information, to the right person, at the right time.

While it is known that the data pre-processing and analytics phases consume most of the effort to be made in the whole data process, there is a remaining effort that must be devoted to the visualization and explanation of the results. One way to tackle this effort is to rely on efficient visualization metaphors and in smart visual interaction paradigms [4].

Regarding visualization, in [5] a survey of visualization technologies tailored for Smart Manufacturing scenarios can be found. Focusing on visual surveillance of all the captured raw data, many proposals rely on the use of a technology to compose observability dashboards[1] that allow experts in the manufacturing process to find relevant information in the data. Those types of technologies also support different data stores that provide SQL-like query language for interacting with stored data. However, many times, domain experts of those Smart Manufacturing scenarios, interested in analyzing data belonging to particular domains, which have specific meanings, have limited programming skills. As a result, they might find it difficult to analyze data that is not visualized in the dashboards.

---

* Corresponding author.
*E-mail addresses:* idoia.berges@ehu.eus (I. Berges), victorjulio.ramirez@ehu.eus (V.J. Ramírez-Durán), a.illarramendi@ehu.eus (A. Illarramendi).

[1] https://logz.io/blog/grafana-vs-kibana/.

Considering that limitation, in this paper we present an alternative that can be complementary to the use of dashboards, and which is based on the use of a semantic-based Visual Query System (VQS) tailored to Industry 4.0 scenarios.

In general, VQSs use a visual representation to depict the domain of interest and provide interaction mechanisms in order to formulate requests to the data stores by means of visual expressions [6].

The VQS that we propose provides a 2D digital representation of a manufacturing machine and dynamically customized forms to formulate queries. In this way domain experts can gain value and insights out of the captured data as rapidly as possible, minimizing the need to contact Information Technology experts. Both the digital representation and the forms are linked with semantic descriptions. The handling of semantic descriptions for data exploration and visualization tasks is what constitutes a novel technical contribution of the proposal. Those semantic descriptions are expressed using terms of different ontologies[2] such as the Semantic Sensor Network Ontology SOSA/SSN [7], the Ontology of units of Measure OM [8], the Semanticscience Integrated Ontology SIO[3] and of one ontology that we have built for an Industry 4.0 scenario, called ExtruOnt [9]: an ontology for describing a type of manufacturing machine. ExtruOnt imports terms from already developed ontologies and incorporates specific terms regarding the components of an extruder machine, their characteristics and spatial connections, and those related to sensors (e.g., `MotorConsumptionSensor`) used to capture observations about the performance of a machine (e.g., `DoubleValueObservation`) in order to favor interoperability issues. Although ExtruOnt focuses on extruders, it is straightforward to consider other types of machines due to the nature of the ontology (see section 3). The semantic descriptions incorporated in the ontology can be stored in a knowledge platform (e.g., Virtuoso). The proposed system provides the following benefits in the data exploration and visualization process:

1. **Possibility of querying the monitored data at a higher level of abstraction**. The system facilitates domain experts to formulate queries by operating on a 2D digital representation of a machine and then customizing them through forms. Those forms are dynamically generated by making use of stored semantic descriptions.
2. **Possibility of downloading semantically enriched data**. Domain experts of smart manufacturing scenarios can download, in an easy way, specific semantic descriptions related to domain data through a form. Using tools they already know (e.g., Excel) they have the possibility to perform new types of analyses with those semantic descriptions that they can not do dealing only with raw data.
3. **Possibility of incorporating on-the-fly semantic annotations in the visualization of results.** The results obtained for the queries are shown using tailored graphical representations customized according to the nature of the data domain. Moreover, they can be enriched with semantic description such as information related to outliers, incorporated to the raw data captured by sensors.

4. **Possibility of providing a customized visualization of a manufacturing machine**. Not all the machines of the same type (e.g., extruders) incorporate the same type of sensors. Visualizing a machine with its specific sensors, and thus, providing customized representations of machines is possible by consulting the semantic descriptions contained in the ontology

The system has been materialized for a real smart manufacturing scenario (in particular, a plastic bottle production factory that follows an extrusion process) and has been tested by domain experts. Thus, as an additional contribution we show the previously mentioned advantages located in this case study together with an empirical evaluation of the usability and of the performance in terms of required space to store the semantic descriptions and queries execution time when dealing with them.

A preliminary version of this work that outlines some basic concepts of our system was presented at the Eighth International Workshop on Modeling and Management of Big Data (MoBiD 2019) [10]. Here, the original workshop paper has been extended to include, among others, a new type of query of greater complexity, a download functionality, two additional visualization options and an empirical evaluation of the performance of the proposal in different storage solutions.

In the rest of this paper we present first some related works. Then, we introduce a background on the semantic descriptions considered in the paper. Next, we present the different resources for data exploration incorporated in the implemented system. Then, we perform an empirical evaluation about the implemented system and its behavior. Finally, we end with some conclusions.

## 2. Related work

In general, visual data explorations allow users to interactively explore the content of the data and identify interesting patterns that may be of their interest, in an autonomous way, without requiring assistance from Information Technology experts. Thus, Key Performance Indicator (KPI) dashboards have been actively used for this purpose in several domains, including manufacturing. For example, in [11] a digital control room that integrates multitouch and multiuser-based annotation dashboards for analyzing manufacturing data is presented. In [12] a calendar view is used to visualize and identify the issues and outliers that occur during the manufacturing process. In [13] a machine learning-based approach is used to support real-time analysis and visualization of sensor and ERP data. What these works have in common is that they directly show the information of a set of previously defined KPIs.

However, visualization systems must also offer customization capabilities to different user-defined exploration scenarios and preferences according to the analysis needs [4]. In this sense, Visual Query Systems already have a track record. They have been used for querying databases [6], for retrieving data from the Web [14] and also for visual exploration of time series. In this last case, there are approaches that advocate for the use of example-based methods such as [15], and [16], which proposes a multilevel map-based visualizations of geolocated time series. Different proposals can also be found among systems that deal with semantic data, such as SparqlFilterFlow [17], which employs a diagram-based approach to represent the queries, and Rhizomer [18], which employs a form based approach. Moreover, OptiqueVQS [19] is a semantic-based visual query system that exploits ontology projection techniques to enable graph-based navigation over an ontology during query construction and sampled data to enhance selection of data values for some data attributes. It shows all the classes defined in the loaded domain ontology to the users as a starting point for queries formulation. This forces domain experts to gather experience in the ontology before using the system. Other works such as

---

[2] Here the term *ontology* refers to the knowledge base composed of the conceptual level (i.e., axioms for classes and properties) and the instance level (i.e., assertions about individuals).

[3] https://bioportal.bioontology.org/ontologies/SIO.

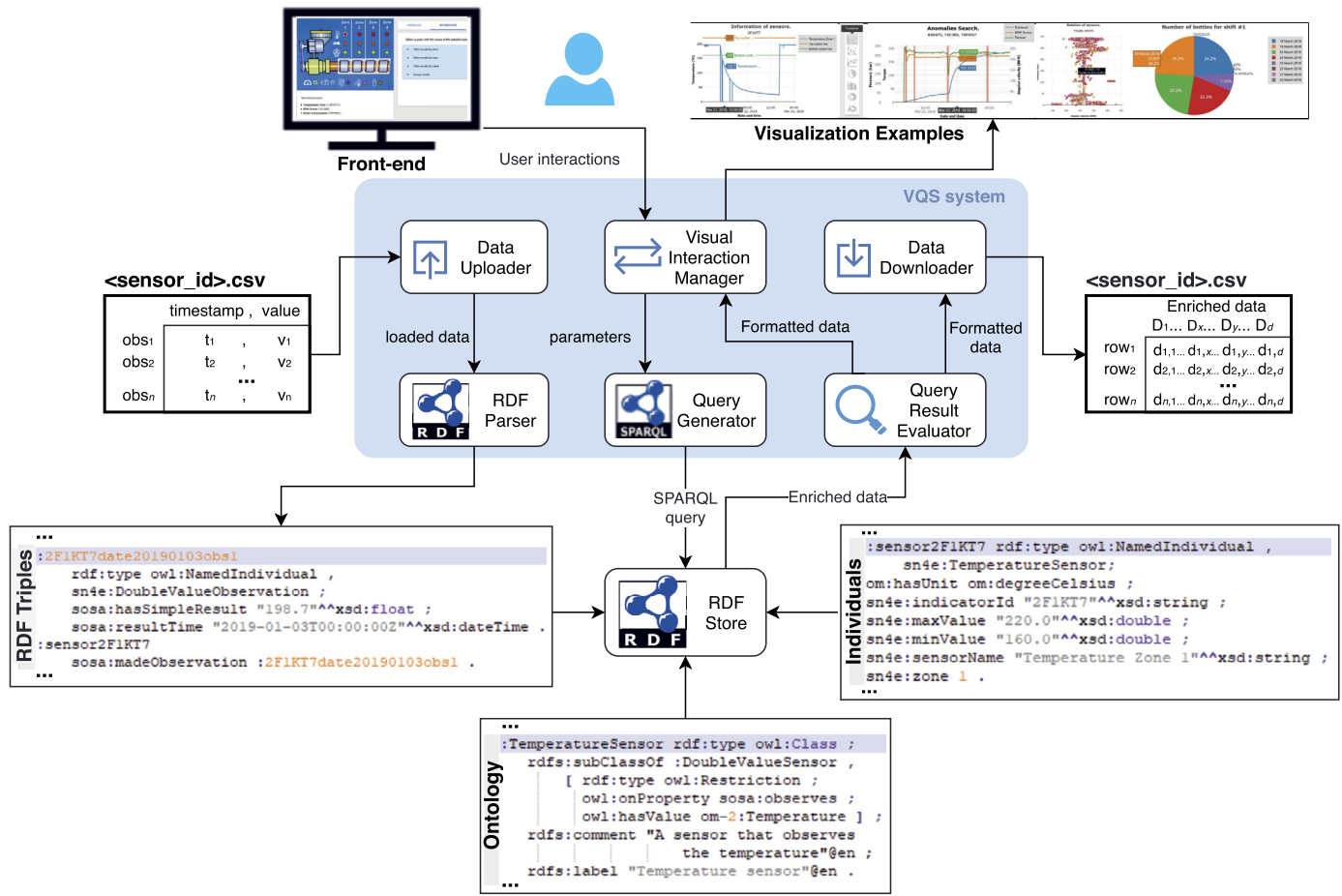

**Fig. 1.** Proposed VQS system overview.

[20] focus on tools such as OWLViz[4] or LODLive [21] for visualizing the content of ontologies and knowledge bases in the form of a graph but not for querying them, or on tools that allow querying but require a good knowledge of the underlying ontology (e.g., [22]).

Few proposals can be found in the literature that provide visual query systems for industrial scenarios and even less that provide semantic-based visual query systems. However, the incorporation of semantic techniques such as ontologies in the manufacturing field is being considered in different proposals due to the benefits that they bring to shift towards a Smart Manufacturing-oriented operation. Thus, ontologies are being defined and used in different manufacturing scenarios ([23,24]). For example, The P-PSO ontology [25] considers four aspects in the manufacturing domain: Product, Physical Aspect, Technological Aspect and Control Aspect. The ontology for Prognostics and Health Management of Machines [26] provides standardization of concepts and terms that are relevant to failure analysis in mechanical components. OntoSTEP [27] allows the description of product information mainly related to geometry.

The semantic based visual query system proposed in this paper (Fig. 1) considers the two aspects mentioned above: lack of visual query systems for smart manufacturing scenarios and interest of incorporating semantics techniques in those scenarios. For this reason, it makes a combined use of a manufacturing-oriented ontology and an interactive visual 2D digital representation of a machine to allow domain experts a smart data exploration pro-cess. The system manages a specific ontology called *ExtruOnt*[5] [9]. Due to a lack of sound descriptions of manufacturing machines that happen to be accessible, interoperable, and reusable we had to develop that ontology for providing detailed descriptions of a real manufacturing machine, in particular, an extruder.

## 3. Background on semantic descriptions

The I4TSMS [28] framework that we are developing aims to provide multiple services in Smart Manufacturing scenarios. Four main software components can be distinguished in its architecture (Fig. 2): The WebApp component, which allocates the Web Application front-end and provides users with a data exploration and visualization artifact; the Pre-Processing component, which is in charge of the captured time-series data preprocessing tasks; the Data and Knowledge Manager, which manages the data and knowledge storage requirements; and finally, the Web Service component, which manages the interconnection between the WebApp component and the Data and Knowledge Manager for providing the answers to the queries formulated by the users.

### 3.1. Ontology description

The Data and Knowledge Manager relies on a knowledge system to manage the semantic descriptions of the main components of the machine and its sensors. Regarding those semantic descriptions, *ExtruOnt* is composed of different modules which contain classes and properties for expressing descriptions about the features of the extruder, such as its components (*components4ExtruOnt*), characteristics (*OM4ExtruOnt*), spatial connections (*spatial4ExtruOnt*) and finally the sensors (*sensors4ExtruOnt*) used to capture indicators about the performance of that type of machine. Of those modules, only the module components4ExtruOnt is dependent on the type of machine. Thus, in order to adapt the approach to other machines, let's say a wire drawing machine, a new main ontology should be created (e.g., WidraOnt), without making major changes on the *OM4ExtruOnt*, *spatial4ExtruOnt*, *sensors4ExtruOnt* and *3D4ExtruOnt* modules (perhaps a new type of sensor, if needed), and only the *components4ExtruOnt* module would need to redefine.

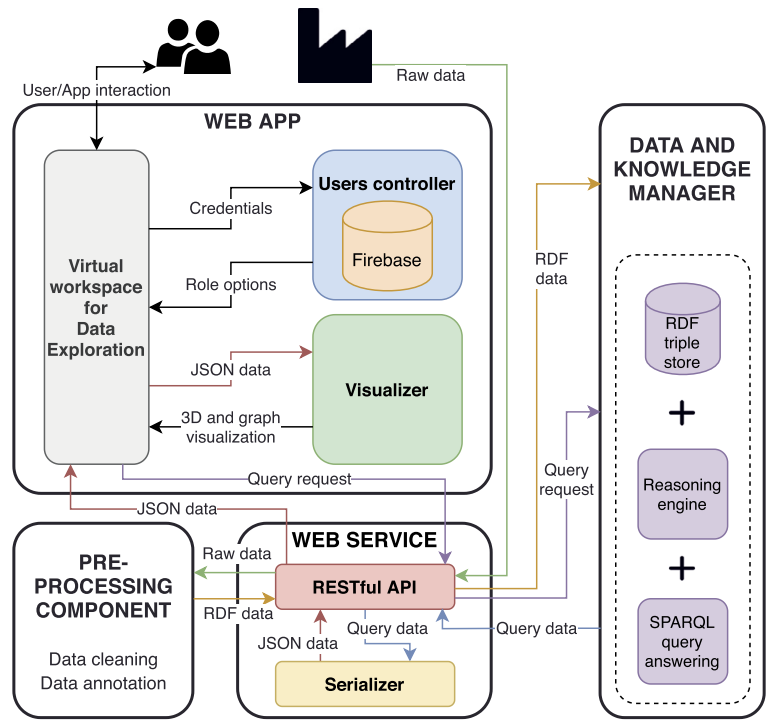

**Fig. 2.** Architecture of the I4TSMS web platform.

In this paper, we mainly focus only on the *sensors4ExtruOnt* module, which contains both specific terms regarding sensors (e.g., `Pressure Sensor`) used to capture observations (e.g., `DoubleValueObservation`), and terms imported from other well-known ontologies, such as SOSA/SSN [7], OM [8] and SIO in order to favor its interoperability. We extend the classes `Sensor`, `Observation` and `ObservableProperty` of the SOSA/SSN ontology to describe more accurately the nuances of the sensors, observations and properties of our domain, and therefore allowing to provide the most suitable graphical representation.

The main class `Sensor` is a specialization of `sosa:Sensor` where two new properties have been included: `indicatorId` (to represent the identifier of the sensor) and `sensorName` (to indicate the name of the sensor). In our scenario, sensors capture either true/false data or numerical data. Thus, two subclasses of `Sensor` have been defined to represent this type of information: `BooleanSensor` and `DoubleValueSensor` respectively. Moreover, these subclasses have been further specialized in order to represent the specific sensors needed to monitor an extruder (see Table 1). Classes `ResistorOnOffSensor` and `FanOnOff-Sensor` are subclasses of `BooleanSensor`, while the remaining classes are subclasses of `DoubleValueSensor`.

For each type of sensor its observable property is indicated by the property `sosa:observes`. For example, the observable property of a `FeedRateSensor` is the `Speed` individual imported from OM. Moreover, for each subclass of `BooleanSensor` the true/false meaning of the observed property can be indicated by properties `meaningState1` and `meaningState0` respectively. For instance, the values of properties `meaningState1` and `meaningState0` in a `ResistorOnOffSensor` are "On" and "Off", which indicates that when the sensor registers the value *true* it means that the sensor is on and when the value is *false* the sensor is off. For each subclass of `DoubleValueSensor`, the expected range of the values captured by these sensors under normal conditions has been specified by properties `min-Value` and `maxValue`. This range will help to identify possible outliers in the captured data. Other properties regarding capabilities of the sensors, such as the feasible measurement range captured by the sensor, are defined using properties from SOSA/SSN. The units of measure have been imported from OM and the graph and chart types from the SIO. Finally, for each type of sensor an axiom has been added which constrains the type of observations made by the sensor through property `sosa:madeObservation`. For each observation its value and timestamp are indicated by properties `sosa:hasSimpleResult` and `sosa:resultTime`. Two subclasses of `Observation` have been defined, depending on the type of value of the observation: `BooleanObservation` and `DoubleValueObservation`. In Fig. 3 an excerpt of the module can be found. For legibility reasons, only the most important properties and classes, and two specific types of sensors (`ResistorOnOffSensor` and `MotorConsumptionSensor`) are fully pictured.

### 3.2. Semantically enriched data

The real data used for testing the performance of the system have been provided by a Capital Equipment Manufacturer (CEM). This company has installed several sensors in the extruders that it manufactures. Those sensors register time-series data with a continuous measurement at 1 Hz (i.e., one measurement per second) of a variety of equipment setting parameters and physical magnitudes (temperatures, pressures, etc.) related to the raw materials,

**Table 1**

Example of the amount of data collected by the sensors of an average extruder machine in one year under normal conditions.

| Sensor type/class | Description | Count | records[a] | Raw[b] | Virtuoso[b] | Stardog[b] | RDFox[c] | Neo4j[b] |
|---|---|---|---|---|---|---|---|---|
| ResistorOnOffSensor | Observes whether a resistor is on or off | 4 | 126,144,048 | 3.41 | 14.79 | 32.87 | 52.89 | 56.32 |
| FanOnOffSensor | Observes whether a fan is on or off | 4 | 126,144,048 | 3.51 | 11.86 | 20.72 | 63.02 | 57.15 |
| TemperatureSensor | Captures the temperature | 10 | 315,360,120 | 9.06 | 32.15 | 51.79 | 157.55 | 145.54 |
| MotorConsumptionSensor | Captures consumption of the motor | 2 | 63,072,024 | 1.73 | 6.25 | 10.36 | 31.51 | 30.17 |
| SpeedSensor | Captures the speed of the rotational parts | 4 | 126,144,048 | 3.51 | 16.33 | 20.72 | 63.02 | 54.13 |
| PressureSensor | Captures the pressure in the extruder | 2 | 63,072,024 | 1.74 | 5.10 | 10.36 | 31.51 | 28.22 |
| MeltingTemperatureSensor | Captures the melting temperature | 2 | 63,072,024 | 1.65 | 7.74 | 10.36 | 31.51 | 27.76 |
| BottlesPerShiftSensor | Captures the number of bottles in a shift | 4 | 126,144,048 | 3.51 | 11.08 | 20.72 | 63.02 | 54.70 |
| Others | Other sensors in the extruder | 19 | 598,774,188 | 16.76 | 66.05 | 98.41 | 299.34 | 270.33 |
| | Total | 51 | 1,607,926,572 | 44.88 | 171.35 | 276.31 | 793.37 | 724.32 |

[a] One record per second.
[b] Disk size in gigabytes.
[c] Memory usage in gigabytes.

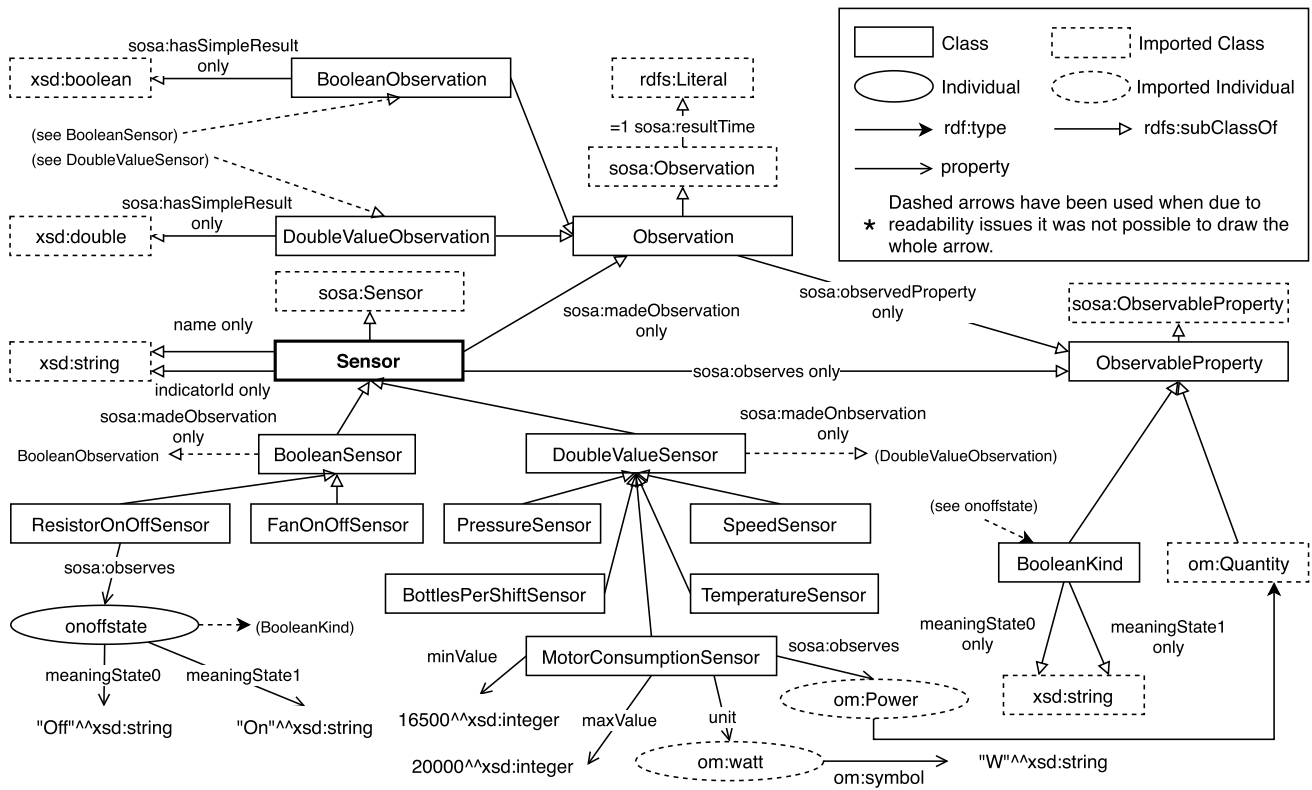

**Fig. 3.** Excerpt of the ontology showing the main classes and properties.

production processes and industrial equipment. In particular, the data came from an extruder machine from a plastic bottles production plant based on an extrusion process, on which the CEM has installed 51 sensors that generate time series of different types (see Table 1 in section 5.2.1). For the tests, a year's worth of data captured by those sensors has been gathered.

In order to provide an enhanced analysis of the data, the raw sensor data is semantically enriched with additional knowledge expressed in an ontology by means of RDF[6] annotations. Although se-mantic technologies were not created for Big Data, we have opted to use them in our proposal, as they provide several advantages such as their ability to describe and integrate heterogeneous data, and to infer new knowledge [29]. The captured data is annotated using the terms in the *ExtruOnt* ontology, converted to a set of RDF triples and stored in the RDF store by means of SPARQL insert queries. The original data contains a pair [value, timestamp] for each of the observations and, for each pair, an instance of Observation is created, indicating the value and its timestamp, along with the type of data, and the sensor. For example let us assume that sensor sensor79PWN7 is an instance of class MotorConsumptionSensor (Fig. 4a) that has made the observa-

```
(a)  :sensor79PWN7 rdf:type :MotorConsumptionSensor .

(b)  :sensor79PWN7 sosa:madeObservation :obs1 .
     :obs1 sosa:hasSimpleResult "18710"^^xsd:double ;
           sosa:resultTime "2018-03-22T19:21:33.559Z"^^xsd:dateTime .

(c)  :obs1 rdf:type :DoubleValueObservation .
     :sensor79PWN7 sosa:observes om:Power ;
                   :minValue "16500"^^xsd:double ;
                   :maxValue "20000"^^xsd:double ;
                   :unit om:watt .
     om:watt om:symbol "W"^^xsd:string .
```

**Fig. 4.** (a) Declaration of `sensor79PWN7` as an instance of `MotorConsumptionSensor`. (b) Example of the triples generated when annotating an observation. (c) Some of the triples that can be inferred.

tion [18710, 2018-03-22T19:21:33.559Z]. Then, the annotations in Fig. 4b are generated, where `obs1` refers to the newly created observation.

Moreover, due to the knowledge available in the ontology, additional information is now related to the observation (Fig. 4c). Since `obs1` was made by `MotorConsumptionSensor` sensor79PWN7, which is a `DoubleValueSensor` that makes `DoubleValueObservations`, `obs1` can be classified as a `DoubleValueObservation`. Furthermore, due to the description of the sensor, it is possible to identify that: `obs1` is an observation of `om:Power`, the value of the observation complies with the expected range of values (between 16500 and 20000) for the observations of this type of sensor, the unit of the observation is `om:watt` and the symbol to represent watts is "W".

## 4. Data exploration

VQSs allow users to analyze data by using a visual interface even if they only have basic technical skills. Thus, VQSs must be both expressive and usable [6].

We have developed an easy-to-use interface but which still allows for performing several types of queries. The user is presented with a dynamically generated picture of the extruder, which consists on a background image of the machine and a top layer where its sensors are placed. The background image is selected depending on the number of zones of the extruder (e.g., 4-zone-extruder, 5-zone-extruder), which can be obtained from the annotations of the machine in the ontology using a SPARQL query. For creating the top layer, another SPARQL query is asked to obtain information about the sensors that are deployed in the machine, along with their type and deployment data. Both SPARQL queries have been implemented within the system and are transparent to the user. Then, clickable bullets representing the sensors, as well as icons that specify their type, are placed in the aforementioned top layer. By using this approach it is possible to provide customized visualizations of multiple extruders. For example, in Fig. 5a the representation of a specific 4-zone-extruder is shown. In this case, 17 sensors of different indicators have been placed dynamically: four `TemperatureSensors`, four `ResistorOnOffSensors`, four `FanOnOffSensors`, a `MotorRPMSensor` (a type of `SpeedSensor`), a `MotorConsumptionSensor`, a `PressureSensor`, a `MeltingTem-peratureSensor` and a `BottlesPerShiftSensor`.

Moreover, at the moment, the system allows for three different types of queries that have been selected in collaboration with domain experts and a download facility that allows one to obtain enriched data enabling thus additional types of analyses. Moreover, due to the modularized nature of the implementation, it could be easily extended to cover other kinds of queries in the future.

### 4.1. Information queries

Information queries are the most simple queries, used to ask for information about the observations of specific sensors. The user selects the sensors by clicking on them and inputs the desired constraints (e.g., date, hour, limits of values, aggregation functions) in a form that is dynamically generated depending on the characteristics of the sensors that have been selected. For example, if sensor 2F1KT7 is selected, the annotations made about it indicate that it is a `TemperatureSensor`, and due to a reasoning process that is also a `DoubleValueSensor`, meaning that it records numerical values. Thus, a slider is shown which allows to restrict the values of the retrieved information to the user's desired range. Moreover, since properties `minValue` and `maxValue` indicate that the usual range for that type of sensor is [160.0, 220.0] and that the unit is `om:degreeCelsius`, the slider has been customized so that values 160.0 °C and 220.0 °C are highlighted (see Fig. 5b), and its limits have been set to the feasible measurement range of the sensor, which in this case is [0.0, 250.0].

Likewise, if sensor URS001 is selected, the annotations indicate that it is a `ResistorOnOffSensor` (and therefore a `BooleanSensor`) and that the true/false values indicate whether the sensor is activated or not. This information is reflected by using an on/off switch. The simplicity of the used design helps users to formulate queries with a high level of abstraction. Once the selection has been made, a SPARQL query is generated and executed against the stored data. Fig. 6(a) shows an example of such a query.

### 4.2. Relation queries

Relation queries are used to ask for the observations made by some specific sensors when certain values hold in the observations made at the same timestamp by some other sensors. First, the user selects all the sensors that take part in the relation. Then they specify which are the sensors whose values they want to ask for, meaning that the remaining selected sensors are the ones whose values are fixed. The user indicates the fixed value for these sensors, which can be a numerical or boolean value (depending on the type of sensor), or the minimum, maximum or average value registered by the sensor in the specified time range. Once again, the form to create the queries is generated dynamically, based on the selected sensors and the information available about them in the ontology.

### 4.3. Anomalies queries

Anomalies queries indicate certain correlations between the values of different sensors that are supposed to hold under normal conditions. The system allows users to run customized or predefined anomalies queries. In order to create a customized anomaly query, after selecting the corresponding sensors the user must establish the correlations between those sensors. For example, one could establish that when the screw rotation speed increases, the pressure and the torque (which is an indicator related to the values of the `MotorConsumptionSensor`) increase as well (see Fig. 5c). Then, this information can be used to locate anomalies in the data (i.e., timestamps where the defined correlations did not hold). The system provides also support for saving the customized anomalies queries as predefined queries so that they can be used in future occasions (see Fig. 5d).

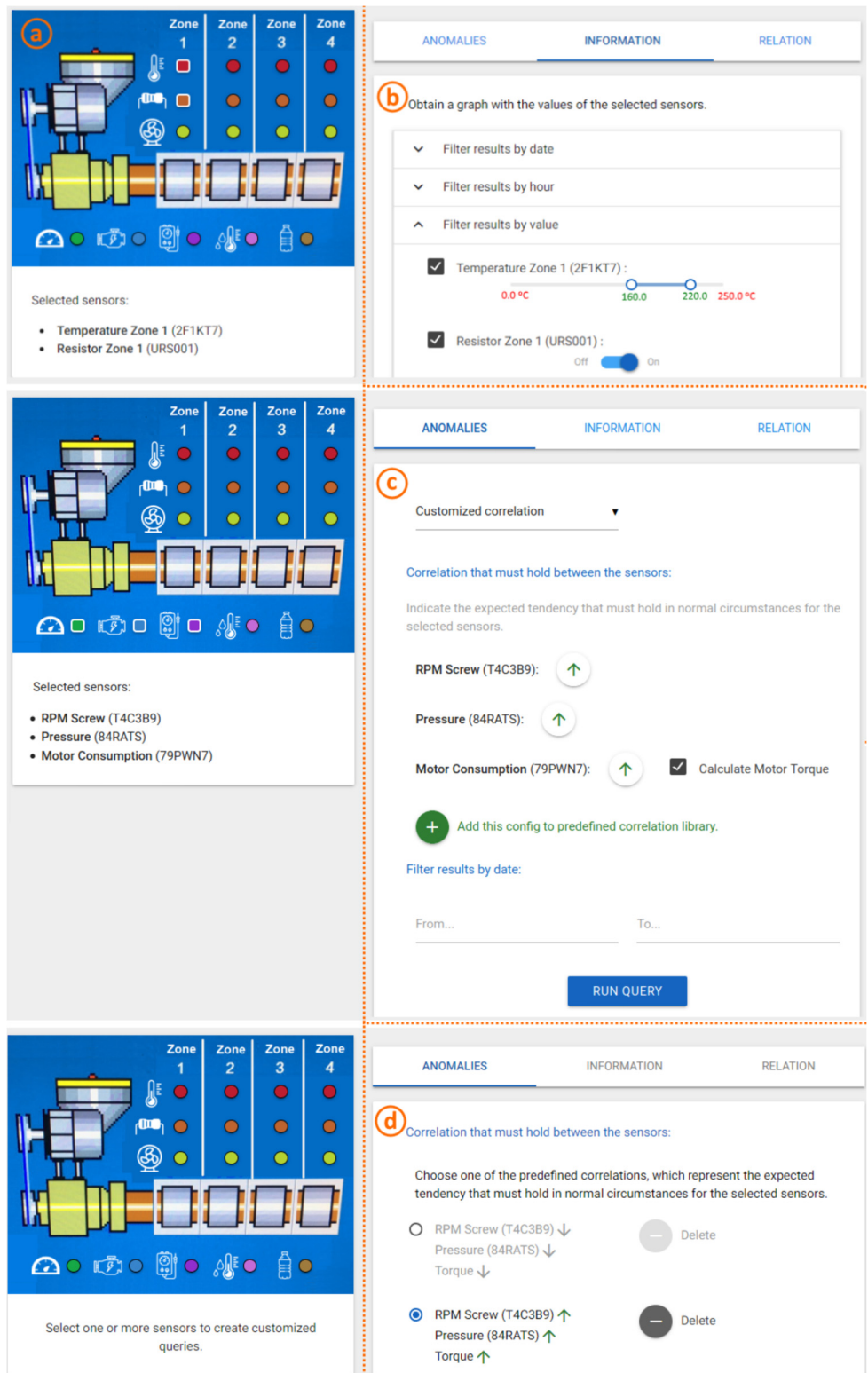

**Fig. 5.** User interface for querying about the data stored in the quad store.

### 4.4. Data visualization

Once a query has been generated, it is executed against the repository of annotated data and the results are shown in form of a chart. In order to select the most suitable representation depending on the nature of the query and the sensors involved, a visualization module has been developed and imported in the sensors module of the ontology, where several recommendations for visualization have been described. As an alternative, an approach to automate data visualization in Big Data analytics but following a model-driven approach can be seen in [30].

```
prefix :<http://bdi.si.ehu.es/bdi/ontologies/ExtruOnt/Instances#>
prefix sosa: <http://www.w3.org/ns/sosa/>
prefix xsd: http://www.w3.org/2001/XMLSchema#
select ?resultValue ?resultTime
where {                                                    (a)
    :sensor2F1KT7 sosa:madeObservation ?obs .
    ?obs sosa:hasSimpleResult ?resultValue ;
        sosa:resultTime ?resultTime .
    filter(?resultValue >= "170"^^xsd:double && ?resultValue <= "200"^^xsd:double) .
    filter((xsd:dateTime(?resultTime) >= "2018-03-20T00:00:00.000Z"^^xsd:dateTime) &&
        (xsd:dateTime(?resultTime) <= "2018-03-22T23:59:59.999Z"^^xsd:dateTime))
} order by asc(?resultTime)

prefix :<http://bdi.si.ehu.es/bdi/ontologies/ExtruOnt/Instances#>
prefix rdfs: < http://www.w3.org/2000/01/rdf-schema#>
prefix owl: < http://www.w3.org/2002/07/owl#>
select ?sensorType ?minValue
where {                                                    (b)
    :sensor2F1KT7 rdf:type ?sensorType .
    ?sensorType rdfs:subClassOf
        [rdf:type owl:Restriction;
                owl:onProperty :minValue ;
                owl:hasValue ?minValue] .
}
```

**Fig. 6.** SPARQL query examples: (a) SPARQL query that is generated when asking for the observations made by sensor 2F1KT7 between 20th and 22nd March 2018 within range 170 and 200. (b) SPARQL query to ask for the minimum value expected for the observations of sensor 2F1KT7.

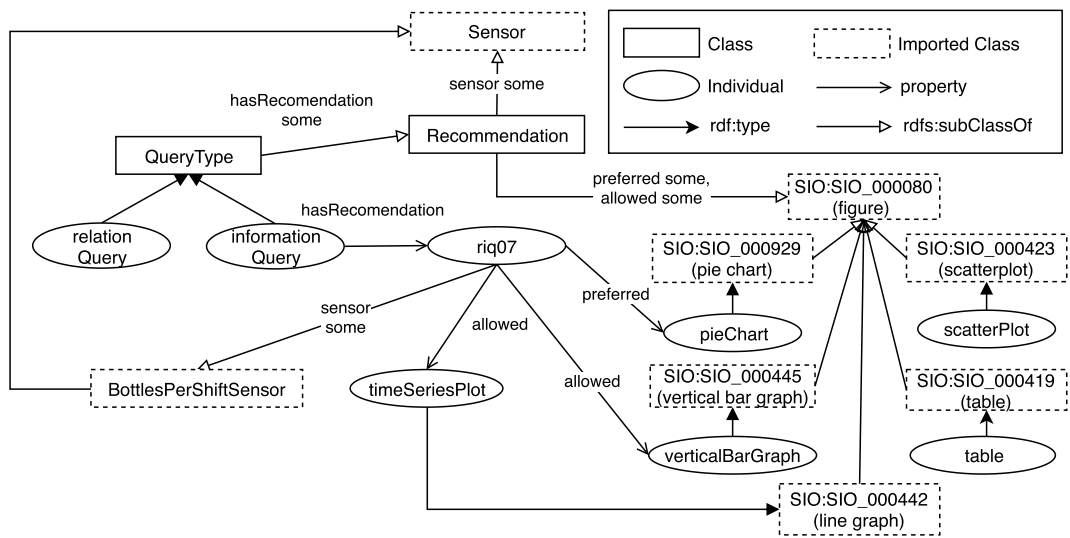

**Fig. 7.** Excerpt of the visualization module of the ontology.

The visualization module uses terms to describe the graph and chart types (e.g., pie chart, bar graph, scatter plot, line graph) from the Semanticscience Integrated Ontology (SIO). This ontology is a simple, integrated upper level ontology for consistent knowledge representation across physical, processual and informational entities. Moreover, we have incorporated additional chart types that were not found in SIO (e.g., gauge chart).

In Fig. 7 an excerpt of the module can be found, with one example of recommendation. More precisely, recommendation `riq07` indicates that when the query is an `information-Query` and the selected sensor is a `BottlesPerShift Sensor`, then the preferred chart is a pie chart (an instance of class `SIO:SIO_000929`), but vertical bar graphs and time-series plots (instances of classes `SIO:SIO_000445` and `SIO:SIO_000442`, respectively) are also allowed.

In Fig. 8a a time-series visualization is used for an information query about temperature sensor 2F1KT7. Top and bottom outlier lines indicate the expected maximum and minimum values for that sensor. The annotations made in the data and the descriptions in the ontology have been used to generate a semantically enriched customization of the chart, as noted in the second benefit explained in section 1. Since sensor 2F1KT7 is a

`TemperatureSensor` that captures values in Celsius degrees, symbol °C is indicated. Moreover, the values for the outlier lines are obtained through SPARQL queries, such as the one in Fig. 6(b) that is used to query about the minimum expected value for that sensor. Fig. 8(b) shows the aforementioned pie chart for a sensor of type `BottlesPerShiftSensor`, indicating the number of bottles and percentage of the total made each day. In Fig. 8(c) a scatterplot has been used to show the relation between the values of two sensors, which measure the screw rotation speed and the pressure. Finally, in Fig. 8(d), a customized time series chart has been used to visualize the results of an `anomaliesQuery` that takes into account the torque, the screw rotation speed and the pressure. A reddish stripe is pictured wherever an anomaly has occurred in the expected trends of the recorded measures. It is also important to say that although our application proposes a visualization of results based on the preferred recommendations in the ontology, then the analyst can select other visualization mode among the allowed ones for that type of query and sensor if it better suits their interest (see Fig. 8(a)).

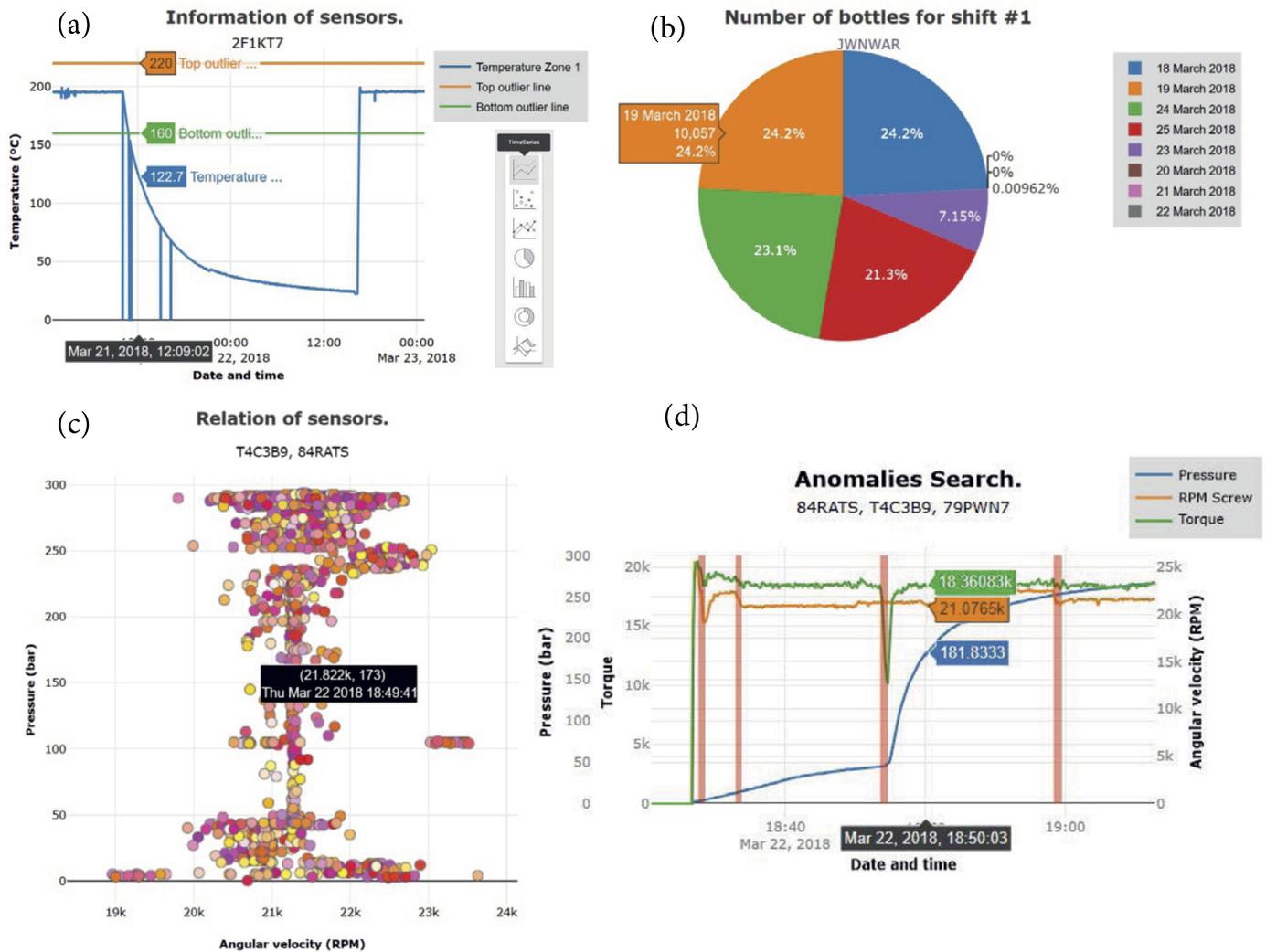

**Fig. 8.** Examples of visualizations of results. (For interpretation of the colors in the figure(s), the reader is referred to the web version of this article.)

### 4.5. Download enriched data

The data download option is divided into several dynamically built sections with which the user is provided not only with the basic data of the sensors (timestamp, value, sensor id, etc.) but also with enriched data (sensor min/max operational values, machine info, data value conversions, etc.) thanks to the semantic descriptions annotated in the underlying ontology. For example, to convert the values of the zone 1 temperature sensor (2F1KT7) from degrees Celsius to Kelvin, the `om:hasFactor` and `om:hasOff-Set` properties of the individual `om:CelsiusScale` belonging to the class `om:IntervalScale` are used, which define the conversion factor (1.0) and the measurement offset (-273.15). In the same way, if a conversion from Kelvin to Fahrenheit is needed, the same properties defined in the individual `om:FahrenheitScale` (`om:hasFactor`=1.8 and `om:hasOff-Set`=-459.67) can be used. The different available scales are related by the dimension to which they correspond (property `om:hasDimension`) and the factor and offset values are based in a specific scale of that dimension. In this way, all the available scales for temperature are grouped in the dimension `om:thermodynamicTemperature-Dimension` and the factor and offset values are based on the scale `om:KelvinScale`. Moreover, those scales are obtained taking into account the unit of measurement set for zone 1 temperature sensor (`om:degreeCelsius`). The definitions of the conversions between measures are annotated by default in the ontology.

Fig. 10 shows the annotations referring to the example described above.

Furthermore, the connections between the different modules of the ontology allow to link the sensors with the components of the extruder where they are located. Likewise, they allow to extract the characteristics related to these components, expanding the amount of enriched data that can be requested by the user. For example, the useful life and performance of the extruder motor in different shifts could be analyzed by comparing its technical specifications (power, speed, torque, etc.) with the values of the motor consumption sensor for the same shifts. All this information can be downloaded directly as shown in the *"Extruder component data related to selected sensors"* section of Fig. 9.

The data download option also includes aggregation functions (average, median, mode, count, sum, etc.) and a filtering by date range. The interface is presented in a simple way where the user selects the types of enriched data that they wish to obtain from a dynamically generated list depending on the sensors selected in the section on the left. Pressing the download button generates a CSV file with the requested information.

## 5. An empirical evaluation

In this section we present an empirical evaluation made in the context of the case study from the points of view of the usability and the behavior.

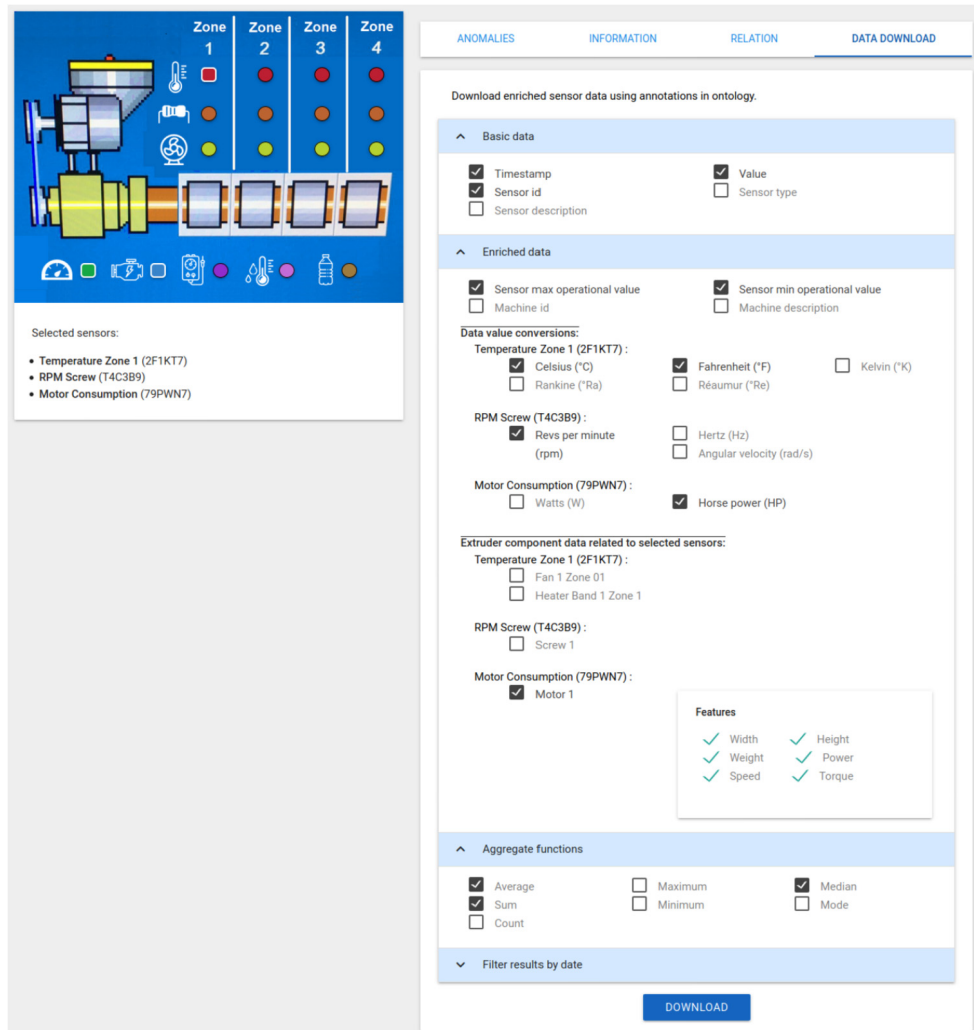

**Fig. 9.** User interface for downloading enriched data.

### 5.1. Evaluation of the usability

A qualitative usability evaluation of the system was performed. It involved two types of persons: 1) A R&D director of a company that develops machines that produce bottles based on an extrusion process, who we work closely with. This person also provides us real data captured from the machines developed by his company. 2) A director of an IBDS (Industrial Big Data Services) Provider company. An IBDS is an ITS (Information Technology Supplier) company that supplies manufacturers with the required technology and services to smartize their manufacturing businesses. Thus, IBDS Providers constitute a fundamental agent in industrial scenarios where there is an interest in adopting Smart Manufacturing approaches.

The users were introduced to the system and we asked them about the following main aspects: how intuitive was the system for performing the queries that they wanted, the grade of customization offered for each of the query types, the elapsed time between running the query and the presentation of the results, and the suitability of the visualization of the results. Moreover, we asked for any other additional suggestions.

They provided positive responses with regard to the intuitiveness of the system for performing queries, the customization of queries that can be achieved through the application, and speed and suitability of the visualization of the results. In addition to some minor changes that are already incorporated in the solution presented in this paper, they made the suggestion that providing the opportunity of choosing between a 3D representation of the extruder and the current 2D representation could enhance the solution, since some users might want to work with a more realistic representation of the machine. We are currently working on that 3D representation.

### 5.2. Evaluation of the behavior

As stated in section 3 the Data and Knowledge Manager must rely on a knowledge system to provide its functionalities. We evaluated the behavior of our solution in three different RDF stores and a NoSQL database. The RDF stores were selected by taking into account the performance results shown in [31] and [32]: Virtuoso[7] (version 08.03.3314), Stardog[8] (Version 7.1.1) and RDFox[9] (Version 2.1.1). The selected NoSQL database was Neo4j[10] (Version 4.1.3), due to its first position in the ranking presented in the DB-Engines website.[11] The hardware setup, where the RDF stores and the NoSQL database were evaluated, consisted of individual instances of m1-ultramem-40 virtual machines (40 Intel Skylake

---

```
:sensor2F1KT7 rdf:type owl:NamedIndividual ,
                    sn4e:TemperatureSensor;
        om:hasUnit om:degreeCelsius ;
        sn4e:indicatorId "2F1KT7"^^xsd:string ;
        sn4e:maxValue "220.0"^^xsd:double ;
        sn4e:minValue "160.0"^^xsd:double ;
        sn4e:sensorName "Temperature Zone 1"^^xsd:string ;
        sn4e:zone 1 .

om:CelsiusScale rdf:type owl:NamedIndividual ,
                        om:IntervalScale ;
        om:hasDimension om:thermodynamicTemperature-Dimension ;
        om:hasScale om:KelvinScale ;
        om:hasUnit om:degreeCelsius ;
        om:hasFactor "1.0"^^xsd:float ;
        om:hasOff-Set "-273.15"^^xsd:float .

om:FahrenheitScale rdf:type owl:NamedIndividual ,
                        om:IntervalScale ;
                om:hasDimension om:thermodynamicTemperature-Dimension ;
                om:hasScale om:KelvinScale ;
                om:hasUnit om:degreeFahrenheit ;
                om:hasFactor "1.8"^^xsd:float ;
                om:hasOff-Set "-459.67"^^xsd:float .

om:RankineScale rdf:type owl:NamedIndividual ,
                        om:RatioScale ;
        om:hasDimension om:thermodynamicTemperature-Dimension ;
        om:hasScale om:KelvinScale ;
        om:hasUnit om:degreeRankine ;
        om:hasFactor "1.8"^^xsd:float ;
        om:hasOff-Set "0.0"^^xsd:float .

om:ReaumurScale rdf:type owl:NamedIndividual ,
                        om:IntervalScale ;
        om:hasDimension om:thermodynamicTemperature-Dimension ;
        om:hasScale om:KelvinScale ;
        om:hasUnit om:degreeReaumur ;
        om:hasFactor "1.0"^^xsd:float ;
        om:hasOff-Set "-218.52"^^xsd:float .
```

**Fig. 10.** Excerpt of the annotations for data value conversions.

virtual CPUs, 961 GB of RAM and SSD with 1200 GB of storage capacity) from the Google Cloud Compute Engine platform. Next some of the evaluation results are presented.

### 5.2.1. Data storage space

Table 1 presents the results of the required storage space for the time-series data captured during one year. With regard to the RDF stores, Virtuoso makes a better space management, decreasing the space needed to store the series by 37.99% and 78.4% compared to Stardog and RDFox, respectively. It is worth mentioning that RDFox is an in-memory RDF triple store, therefore, it uses the Random Access Memory to store the data instead of disk space. This feature makes the data loading faster than the other tested RDF stores but penalizes the amount of memory used due to poor compression.

In the case of the NoSQL database Neo4j, the storage space management does not outperform Virtuoso either. It had been selected for the empirical evaluation because it also supports the concept of relationship. In this way, it is possible to represent an RDF triple as the relationship between two nodes. However, the way in which the RDF triples were loaded into Neo4j differs from the way used for RDF stores. While SPARQL inserts are used for the latter, the Neo4j RDF & Semantics toolkit (n10s[12]) is used for the former. n10s is a plugin that enables the use of RDF in Neo4j and can be used to import existing RDF datasets, build integrations with RDF generating endpoints or easily construct RDF endpoints on Neo4j.

The main drawback in Neo4j is that there not exists an underlying ontology and after the transformation of the RDF data into graph data, all the RDF, RDFS, and OWL tags lose their semantic meaning becoming just simple labels, preventing the use of reasoning with them.

### 5.2.2. Query response times

Fig. 11 shows the response time for the three types of queries presented in section 4.

**Information Queries.** The query response time evaluation for information queries is shown in Fig. 11(a) using the three different RDF stores named previously and three different time windows. The queries were executed 10 times each and the average value was calculated. As it can be seen, on the one hand, Virtuoso is the fastest RDF store for the three different time windows, even faster than the in-memory RDFox store. On the other hand, Stardog presents a low performance solving this type of queries with a considerable distance from the others.

**Relation Queries.** In Fig. 11(b) an evaluation of the query response time for this type of queries is presented. RDFox is the fastest RDF store for those queries with a time window of a day and a week. However, Virtuoso presents a better performance for those relation queries with a longer time window. Also, the query response time in Virtuoso for shorter time windows remains constant (4.9 seconds for a day and a week time window). Stardog continues to show poor performance with very high query response times.

**Anomalies Queries.** This type of query demands a different approach with respect to the other ones. In this case, it is necessary to ask about the behavior of the values over time, detecting increases and decreases in those values. In time series, the observations are taken with a certain periodicity, and the values between one observation and the next may fluctuate depending on the sensitivity, precision and tolerance of the sensor, in addition to the noise generated by external factors. Thus, to correctly detect increases and decreases, it is necessary to establish a minimum variation in the values of the observations for a period of time, so that these increases and decreases are not the product of errors generated by the nature of the sensor or external factors. These minimal variations depend on the kind of observation to be evaluated. For example, the power consumption of a motor can vary between 16,500 and 20,000 watts in contrast with the temperature of an extruder band heater, which can vary between 160 and 220 degrees Celsius. This means that oscillations are greater in the observations of the power consumption than in the temperature, needing to define independent minimum variations for each type of observation.

---

[12] https://neo4j.com/labs/neosemantics/.

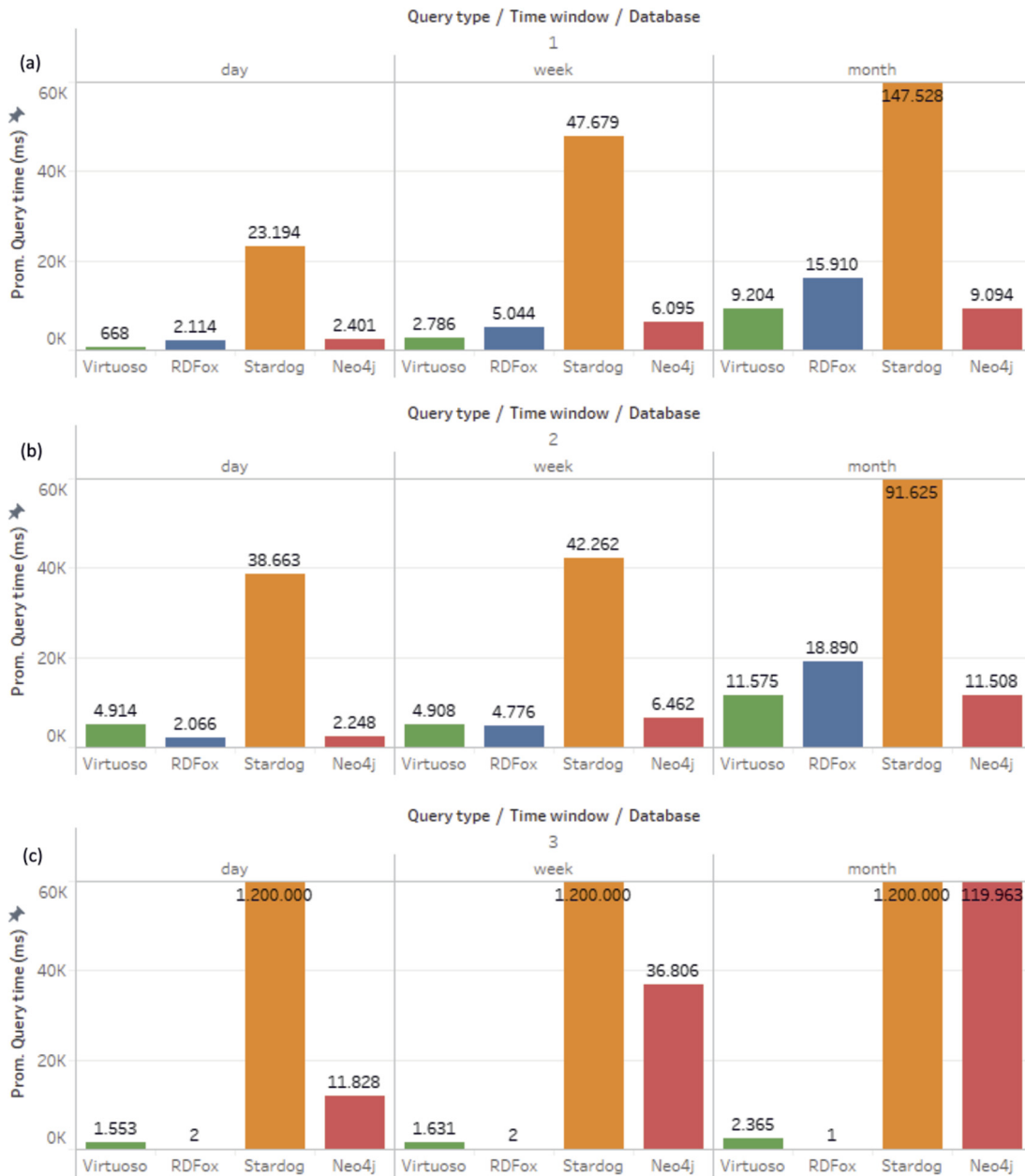

**Fig. 11.** Query response time for: (a) information queries; (b) relation queries; (c) anomalies queries.

The detection of increases and decreases in values can be done by applying rules, which is one of the advantages provided by RDF stores, creating an additional RDF triple to indicate whether in a period of time there is an increase or decrease in value and limiting the query to ask only for the behavior of those triples. However, each RDF store supports a specific rule language and evaluation method, i.e., RDFox supports the use of Datalog rules with the materialization of inferred triples; Virtuoso supports the creation of custom inference rules using the SPIN vocabulary; and Stardog supports two different syntaxes for defining rules, the first is native Stardog Rules syntax (based on SPARQL) and the second is the standard RDF/XML syntax for SWRL. Virtuoso and Stardog use the query rewriting method to evaluate the rules. Fig. 12 shows an example of the syntax used for a rule in RDFox.

In the query response time evaluation presented in Fig. 11(c), it can be seen that the materialization method for rule application used in RDFox provides a remarkable performance compared to Virtuoso and Stardog. Nevertheless, Virtuoso shows a notewor-

```
demo:valueIncrement[?sensor,?actDate] :-
  eo:MotorRPMSensor[?sensor], sosa:madeObservation[?sensor,?obs1],
  sosa:resultTime[?obs1,?actDate], demo:hasAVGResult[?obs1,?actVal],
  BIND (?actDate + "-PT15M"^^xsd:duration AS ?pasDate),
  sosa:madeObservation[?sensor,?obs2],sosa:resultTime[?obs2,?pasDate],
  demo:hasAVGResult[?obs2,?pasVal], FILTER(?actVal > ?pasVal+500)  .
```

**Fig. 12.** Datalog rule to materialize an increase in the value of observations for all motor RPM sensors when the increase is greater than 500 units in a time span of 15 minutes.

thy performance even using the query rewriting method. Finally, queries made in Stardog exceeded the maximum waiting time of two minutes.

Regarding the performance of Neo4j to solve the three types of defined queries, it can be observed that the response times for information and relation queries are similar to those obtained with the best performing RDF stores (i.e., Virtuoso and RDFox). However, for anomaly queries, the query response time increases to about 2 minutes as, to the best of our knowledge, it is not possible to use rules in Neo4j. Therefore, it is necessary to first query the basic data and then apply some post-processing using an external framework (e.g., Spark,[13] Hadoop[14]) to detect anomalies.

## 6. Conclusion

The development of software tools that support customization capabilities that facilitate data exploration and visualization, to different users according to their analysis needs, is a challenge that is being considered in manufacturing scenarios. Exploration and visualization of captured time-series data provides increasing knowledge about the indicators used in the monitoring of machines. In this paper we have presented a semantic-based visual query system that enables domain experts to formulate queries dealing with a customized digital representation of the machine and on-the-fly generated forms. The system also offers the capability of downloading enriched data. Moreover, it provides a tailor-made visualization of the results depending on their nature. The whole process is supported by an underlying ontology where the main components of the machine and its sensors have been described.

Although disk space is a crucial concern when referring to big data scenarios, the benefits of semantic data annotation for data analysis purposes are not comparable with the limited knowledge extracted directly from raw data. Therefore, the increase in storage due to semantic annotation is acceptable and moreover, the query response times obtained are manageable.

## Declaration of competing interest

The authors declare that they have no known competing financial interests or personal relationships that could have appeared to influence the work reported in this paper.

## Acknowledgements

The authors would like to thank Urola Solutions for their help with information about the extrusion process and for providing real data. This research was funded by the Spanish Ministry of Economy and Competitiveness under Grant No. FEDER/TIN2016-78011-C4-2-R and the Basque Government under Grant No. IT1330-19. The work of Víctor Julio Ramírez is funded by the Spanish Ministry of Economy and Competitiveness under contract with reference BES-2017-081193.

---

[13] https://spark.apache.org/.
[14] https://hadoop.apache.org/.